\documentclass{article}
\PassOptionsToPackage{numbers, compress}{natbib}

\usepackage[preprint]{neurips_2024}

\usepackage[utf8]{inputenc}%

\usepackage{blindtext}
\usepackage{fancyhdr}

\usepackage[textsize=tiny]{todonotes}

\usepackage[utf8]{inputenc} %
\usepackage[T1]{fontenc}    %
\usepackage{hyperref}       %

\usepackage{url}            %
\usepackage{booktabs}       %
\usepackage{amsfonts}       %
\usepackage{nicefrac}       %
\usepackage{microtype}      %
\usepackage{xcolor}         %

\usepackage{pifont}

\usepackage{amsmath,amsthm}
\usepackage{amsfonts}
\usepackage{multicol}
\usepackage{multirow}
\usepackage{mathtools}
\usepackage{bm} %
\usepackage{bmpsize}
\usepackage{xcolor}
\usepackage{lipsum}
\usepackage{textcomp}
\usepackage{bm}
\usepackage{gensymb}
\usepackage{nicefrac}
\usepackage{bbm}
\usepackage{tabularx}
\newcolumntype{b}{>{\centering\arraybackslash}X}
\newcolumntype{s}{>{\hsize=.5\hsize}X}
\usepackage{float}
\usepackage{pdfpages}
\usepackage{subcaption, booktabs}
\usepackage{graphicx}
\usepackage{tikz}
\usepackage{xspace}
\usetikzlibrary{automata, positioning, shapes.arrows, chains, math}
\usepackage{todonotes}
\newtheorem{Assumption}{Assumption}

\usepackage{wrapfig}

\usepackage{siunitx}
\usepackage{booktabs}
\usepackage{multirow}
\usepackage{colortbl}
\definecolor{darkgreen}{RGB}{0,100,0}
\newcommand{\red}[1]{{\color{red}#1}}
\newcommand{\green}[1]{{\color{darkgreen}#1}}

\DeclareMathOperator*{\argmax}{arg\,max}

\newif\ifdraft
\drafttrue

\usepackage{booktabs,arydshln}
\makeatletter
\def\adl@drawiv#1#2#3{%
        \hskip.5\tabcolsep
        \xleaders#3{#2.5\@tempdimb #1{1}#2.5\@tempdimb}%
                #2\z@ plus1fil minus1fil\relax
        \hskip.5\tabcolsep}
\newcommand{\cdashlinelr}[1]{%
  \noalign{\vskip\aboverulesep
           \global\let\@dashdrawstore\adl@draw
           \global\let\adl@draw\adl@drawiv}
  \cdashline{#1}
  \noalign{\global\let\adl@draw\@dashdrawstore
           \vskip\belowrulesep}}
\makeatother

\makeatletter
\DeclareRobustCommand\onedot{\futurelet\@let@token\@onedot}
\def\@onedot{\ifx\@let@token.\else.\null\fi\xspace}

\usepackage{pifont}

\usepackage{listings}

\usepackage{xparse}

\NewDocumentCommand{\code}{v}{%
\texttt{\textcolor{black}{#1}}%
}

\ifdraft
\newcommand{\franzi}[1]{\textcolor{purple}{F: #1}}
\newcommand{\adam}[1]{\textcolor{red}{Adam: #1}}
\newcommand{\chris}[1]{\textcolor{green}{C: #1}}
\newcommand{\roy}[1]{\textcolor{blue}{Roy: #1}}
\newcommand{\emmy}[1]{\textcolor{orange}{Emmy: #1}}
\newcommand{\stephan}[1]{\textcolor{teal}{[Stephan: #1]}}
\newcommand{\nicolas}[1]{\textcolor{red}{N: #1}}
\newcommand{\anvith}[1]{\textcolor{brown}{anvith: #1}}
\newcommand{\patty}[1]{\textcolor{olive}{patty: #1}}

\else
\newcommand{\franzi}[1]{}
\newcommand{\roy}[1]{}
\newcommand{\chris}[1]{}
\newcommand{\adam}[1]{}
\newcommand{\emmy}[1]{}
\newcommand{\stephan}[1]{}
\newcommand{\nicolas}[1]{}
\newcommand{\anvith}[1]{}
\newcommand{\patty}[1]{}

\fi

\usepackage{ifthen}
\newboolean{review}  
\setboolean{review}{false}
\ifthenelse{\boolean{review}}{

\newcommand{\out}[1][]{\textcolor{blue}{[...]}}
}
{

\newcommand{\out}[1][]{\textcolor{blue}{}}
}
\definecolor{mygray}{gray}{0.4}
\newcommand{\cmark}{\color{mygray}\ding{51}}%
\newcommand{\xmark}{\color{mygray}\ding{55}}%
\newcommand{\rcmark}{\color{red}\ding{51}}%
\usepackage{algcompatible}
\usepackage[ruled,vlined]{algorithm2e}

\algdef{SE}[SUBALG]{Indent}{EndIndent}{}{\algorithmicend\ }%
\algtext*{Indent}
\algtext*{EndIndent}

\SetKwInput{KwInput}{Input}                %
\SetKwInput{KwOutput}{Output}              %

\usepackage[capitalize,noabbrev]{cleveref}
\theoremstyle{plain}
\newtheorem{theorem}{Theorem}[section]

\theoremstyle{definition}
\newtheorem{definition}[theorem]{Definition}

\theoremstyle{remark}
\newtheorem{remark}[theorem]{Remark}
\title{BACON: Bayesian Optimal Condensation Framework for Dataset Distillation}

\author{%
Zheng Zhou\textsuperscript{$\dagger$},\;
Hongbo Zhao\textsuperscript{$\dagger$},\;
Guangliang Cheng\textsuperscript{$\diamondsuit$},\;
Xiangtai Li\textsuperscript{$\spadesuit$},\;
Shuchang Lyu\textsuperscript{$\dagger$}\thanks{Corresponding authors},\;\\
\textbf{Wenquan Feng}\textsuperscript{$\dagger$}\textbf{,}\;
\textbf{Qi Zhao}\textsuperscript{$\dagger$}\\
\textsuperscript{$\dagger$}School of Electronic and Information Engineering, Beihang University\\
\textsuperscript{$\diamondsuit$}Department of Computer Science, University of Liverpool\\
\textsuperscript{$\spadesuit$}S-Lab, Nanyang Technological University\\
\small \texttt{\{zhengzhou, bhzhb, lyushuchang, buaafwq, zhaoqi\}@buaa.edu.cn}\\
\small \texttt{Guangliang.Cheng@liverpool.ac.uk}\\
\small \texttt{xiangtai.li@ntu.edu.sg}
}

\begin{document}

\maketitle

\begin{abstract}

  Dataset Distillation (DD) aims to distill knowledge from extensive datasets into more compact ones while preserving performance on the test set, thereby reducing storage costs and training expenses. However, existing methods often suffer from computational intensity, particularly exhibiting suboptimal performance with large dataset sizes due to the lack of a robust theoretical framework for analyzing the DD problem. To address these challenges, we propose the \underline{\textbf{BA}}yesian optimal \underline{\textbf{CON}}densation framework (\underline{\textbf{BACON}}), which is the first work to introduce the Bayesian theoretical framework to the literature of DD. This framework provides theoretical support for enhancing the performance of DD. Furthermore, BACON formulates the DD problem as the minimization of the expected risk function in joint probability distributions using the Bayesian framework. Additionally, by analyzing the expected risk function for optimal condensation, we derive a numerically feasible lower bound based on specific assumptions, providing an approximate solution for BACON. We validate BACON across several datasets, demonstrating its superior performance compared to existing state-of-the-art methods. For instance, under the IPC-10 setting, BACON achieves a 3.46\% accuracy gain over the IDM method on the CIFAR-10 dataset and a 3.10\% gain on the TinyImageNet dataset. Our extensive experiments confirm the effectiveness of BACON and its seamless integration with existing methods, thereby enhancing their performance for the DD task. Code and distilled datasets are available at \href{https://github.com/zhouzhengqd/BACON}{BACON}.
\end{abstract}

\vspace{-0.2in}
\section{Introduction}
\label{sec:intro}
\begin{figure}
	\centering
	\includegraphics[width=1\textwidth]{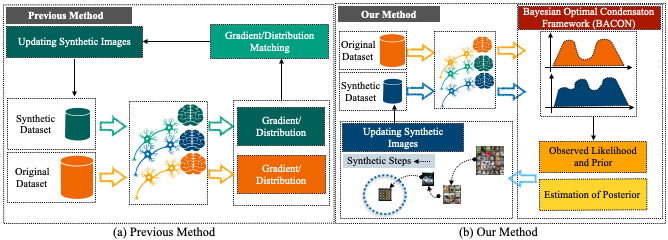}
	\caption{\textbf{Comparison between our method and previous methods:} (a) Existing state-of-the-art DD methods typically rely on a common paradigm involving the alignment of gradients \cite{b28} and distributions \cite{b11, b56} computed by neural networks on both original and synthetic datasets. (b) In contrast, our BACON method transfers the DD task into the Bayesian optimization problem and generates synthetic images by assessing the likelihood and prior probabilities.}
	\label{fig1}

\end{figure}
Dataset Distillation (DD) is an emerging research topic focused on distilling a large dataset into a smaller set of synthetic samples \cite{b22}. This process enables models trained with these synthetic samples to achieve performance comparable to those trained on the entire dataset. DD is typically framed as a meta-learning problem \cite{b23} involving bilevel optimization. In the inner-loop optimization, the learnable parameters of neural networks are trained, while in the outer-loop optimization, synthetic samples are generated by minimizing the classification loss on the original samples. Subsequently, Zhao {\it et al.} \cite{b28} proposed Dataset Condensation (DC) to enhance the efficiency of DD tasks through gradient matching, as illustrated in Figure~\ref{fig1}(a). Since 2018, DD has undergone significant development, leading to the emergence of various approaches. These include performance-matching methods (e.g., DD \cite{b22}, KIP \cite{b24, b25}, RFAD \cite{b26}, and FRePo \cite{b27}), parameter-matching methods (e.g., DC \cite{b28}, MTT \cite{b29}, TESLA \cite{b30}, and FTD \cite{b31}), and distribution-matching methods (e.g., DM \cite{b11}, CAFE \cite{b32}, and IDM \cite{b56}) along with other approaches \cite{b33,b34,b36,b37,b39,b50,b51,b57,b62}.

The development of DD and DC has greatly facilitated the advancement of DL and spurred innovation in various downstream tasks, including continual learning \cite{b40,b41,b42}, federated learning \cite{b43,b44,b45,b46}, knowledge distillation \cite{b47}, and adversarial learning \cite{b63,b64,b65,b66,b67}. However, existing methods often face computational intensity and demonstrate suboptimal performance with large datasets due to the absence of a robust theoretical framework for analyzing the optimization problem in DD. To explore potential solutions to these challenges, we propose the following key questions: 
\begin{itemize}
  \item[*] {\it How can we effectively formulate the DD problem?}
  \item[*] {\it What is the theoretical lower bound of optimal condensation?}
\end{itemize}
To address the above questions, we propose the \underline{\textbf{BA}}yesian optimal \underline{\textbf{CON}}densation framework (\underline{\textbf{BACON}}), which is the first work to introduce the Bayesian theoretical framework to the DD field. BACON provides a sound theoretical analysis framework for DD tasks, enabling us to formulate these tasks as the minimization of the expected risk function within joint probability distributions and to derive the theoretical lower bound of the risk function. In particular, we present a theoretical framework that facilitates the measurement of the risk of expectations in joint probability distributions. This framework leverages the output of a neural network trained on both original and synthetic datasets to represent the probability distribution. Subsequently, we analyze the optimal condensation risk function to solve the DD problem and obtain the approximated solution for BACON based on certain assumptions. Lastly, we design a highly effective optimization strategy for the DD task using BACON. The entire process of BACON is depicted in Figure~\ref{fig1}(b). We validate the proposed BACON on multiple image classification benchmarks and demonstrate its significant superiority over the state-of-the-art methods (DD \cite{b22}, LD \cite{b57}, DC \cite{b28}, DSA \cite{b33}, DCC \cite{b62}, CAFE \cite{b32}, DM \cite{b11} and IDM \cite{b56}) with multiple datasets such as MNIST \cite{b58}, Fashion-MNIST (F-MNIST) \cite{b60}, SVHN \cite{b61}, CIFAR-10 \cite{b52}, CIFAR-100 \cite{b52} and TinyImageNet \cite{b53}. Furthermore, we assess the efficacy of various components and hyperparameters of BACON through a series of ablation studies. Finally, we visualize the synthetic image generated by BACON under different settings.

Our \textbf{contributions} can be summarized as follows:
\begin{itemize}
    \item To the best of our knowledge, we are the {\it \textbf{first}} to introduce the Bayesian theoretical framework to the DD task, providing the theoretical support for improving distillation performance.
    \item We present the BACON, a novel and efficient method for the DD task. BACON utilizes the Bayesian framework to formulate the DD problem as the minimization of expected risk function in joint probability distributions. 
    \item Through comprehensive analysis, we derive a numerically feasible lower bound for minimizing the expected risk function in joint probability distributions, based on certain assumptions.
    \item Experimental results demonstrate the superiority of BACON over existing approaches, which can be seamlessly integrated as a plug-and-play module into existing methods.
\end{itemize}
\section{Related Work}
\label{sec:related work}
\paragraph{Dataset Pruning} The traditional approach for reducing training dataset sizes is dataset pruning, also referred to as core-set selection \cite{b18}. This method aims to gather the most representative and valuable samples from the original dataset, resulting in a smaller yet comparable dataset without compromising model performance. While various techniques such as Herding \cite{b19}, K-center \cite{b20}, and the Forgetting method \cite{b21} have been explored, they suffer from three main drawbacks: 1) reliance on heuristic algorithms, 2) prone to local optima, and 3) the omission of many sub-optimal representative samples due to employed dropping strategies.
\paragraph{Dataset Distillation (DD)} Unlike core-set selection, DD achieves optimal performance with complete representative features. DD was first proposed by Wang {\it et al.} \cite{b22} as a bi-level optimization problem. However, tackling the bi-level optimization problem entails additional computational expenses due to its nested recursion. To mitigate these overheads, Zhao {\it et al.} \cite{b28} introduced a gradient matching method called Dataset Condensation (DC). This method enhances overall performance by matching the informative gradients calculated from the original datasets with those from the synthetic datasets at each iteration. Furthermore, Zhao {\it et al.} \cite{b11} presented a distribution matching method referred to as DM, employing the Maximum Mean Discrepancy (MMD) measurement metric. To further improve the distillation performance, Zhao {\it et al.} \cite{b56} proposed a more efficient and promising method known as Improved Distribution Matching (IDM), built upon the DM. In terms of the research on theoretical foundations for DD, Shang {\it et al.} \cite{b48} were the first to introduce information theory into DD research. They formulated DD as a mutual information maximization problem within the information theory framework. Subsequent studies have explored various optimization objectives to constrain image synthesis, such as DSA \cite{b33}, CAFE \cite{b32}, MTT \cite{b29}, DREAM \cite{b51} and others \cite{b34,b36,b37,b39,b50,b51}. 
\paragraph{Bayesian Framework for Matching Gradients Method} The Bayesian framework for matching gradients method was first proposed by Mislav {\it et al.} \cite{b49} in the field of gradient leakage, which aimed to reconstruct datasets by aligning gradients. They formalized the problem of gradient leakage as the Bayes optimal adversary framed as an optimization problem to achieve higher reconstruction success.

In contrast, we introduce a novel and optimal condensation framework based on Bayesian optimization for the DD task. Inspired by the incorporation of gradient leakage \cite{b49}, DM \cite{b11}, and IDM \cite{b56}, our approach diverges significantly from previous methods. Unlike the work of Mislav {\it et al.} \cite{b49}, which aligns gradients with respect to the weights of the neural network, we align the joint probability distribution of the neural network output to generate synthetic samples. Furthermore, whereas DM and IDM \cite{b11, b56} focus merely on distribution matching, our method, BACON, leverages the Bayesian framework to comprehensively analyze the expected risk function of probabilities. This approach allows us to derive the theoretical lower bound of risk, effectively bridging the gap between theory and practice under certain assumptions. Further details are provided in Appendix \ref{app:extended background}.

\section{Bayesian Optimal Condensation Framework}
\label{sec:method}
\begin{figure*}
	\centering
	\includegraphics[width=0.92\linewidth]{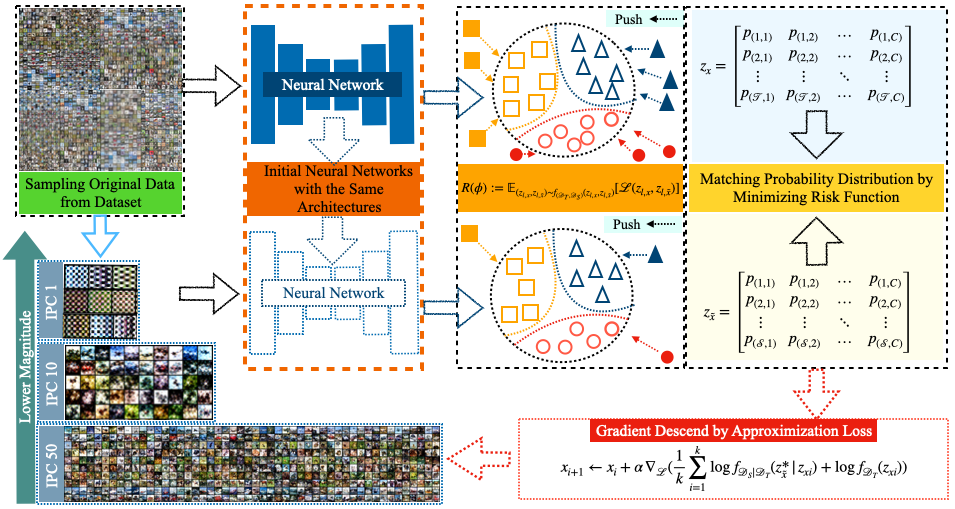}
	\caption{\textbf{The framework of proposed BACON:} The neural networks output the distribution after processing both synthetic and real datasets. Subsequently, we formulate the distribution between the synthetic dataset and real datasets as the Bayesian optimal condensation risk function (refer to Section~\ref{sec:3.2}). The optimal solution of the risk function is derived using the Bayesian formula (refer to Section~\ref{sec:3.3}). To obtain the approximated solution of BACON, we introduce two assumptions (refer to Section~\ref{sec:3.4}), and outline the entire algorithm of BACON in Algorithm~\ref{alg1} (refer to Section~\ref{sec:3.5}).}
	\label{fig2}
  \end{figure*}
In this section, we present the motivation behind our exploration of BACON. We define the Bayesian optimal condensation risk function based on the probability distribution and provide a brief introduction to the meaning of notations. Subsequently, we derive the risk function and determine the theoretically optimal solution. Finally, we propose assumptions for the log-likelihood and prior probability to obtain approximate solutions, aiming to streamline practical implementation and delineate the training strategy. Additional details regarding our method are provided in Appendix \ref{app:proofs}, which includes proofs and discussions.
\subsection{Motivation}
Dataset Distillation (DD) is crucial for reducing storage costs and training expenses while maintaining test performance. However, current methods often face significant computational challenges, especially with large datasets, due to the lack of a robust theoretical framework. We identified two key questions to address these challenges: 1) {\it How can we effectively formulate the DD problem?} and 2) {\it What is the theoretical lower bound of optimal condensation?} Inspired by Mislav {\it et al.} \cite{b49}, who used the Bayesian optimal adversary framework to address gradient leakage, we extend this approach to DD to find the optimal path for maximizing distillation performance while maintaining test accuracy. Subsequently, we propose BACON, a novel method that provides the first theoretical analysis of optimal condensation using Bayesian principles. This robust theoretical foundation significantly enhances distillation performance. The training strategy of BACON is illustrated in Figure~\ref{fig2}.
\subsection{Expected Risk Function in Joint Probability Distribution} 
\label{sec:3.2}
\begin{definition}[Similarity Indicator of $\epsilon$-neighborhood]
	\label{d1}
	Let $z_x$ and $z_{\tilde{x}}$ be two distributions. A binary loss function $\mathcal{L}$ is used to assess their similarity. When the Euclidean distance between the two distributions is small, $\mathcal{L}$ is equal to $0$; otherwise, $\mathcal{L}$ is equal to $1$. As $\epsilon$ approaches zero, the synthetic distribution $z_{\tilde{x}}$ approximates the original distribution $z_x$. We define the similarity indicator function as follows:
	\begin{equation}
		\mathcal{L}(z_x,z_{\tilde{x}}) := \mathbbm{1}\{\Vert z_x - z_{\tilde{x}} \Vert_2 \geq \epsilon\}.
	\end{equation}
\end{definition}
\begin{definition}[Expected Risk Function]
	\label{d2}
	Consider a joint probability distribution $p(z_{x},z_{\tilde{x}})$ formed by the probability distribution of the outputs of neural networks, $\mathcal{D}_T$ and $\mathcal{D}_S$, for the original dataset $\mathcal{T}$ and the synthetic dataset $\mathcal{S}$. Given $z_{x} \in \mathcal{D}_T$ and $z_{{\tilde{x}}}\in \mathcal{D}_S$, the expected risk function in the joint probability distribution $R(\phi)$ is defined as follows:
	\begin{equation}
		R(\phi) := \mathbb{E}_{(z_{x},z_{\tilde{x}}) \sim p(z_{x},z_{\tilde{x}})}[\mathcal{L}(z_{x},z_{ \tilde{x}})],
	\end{equation}
	where $z_{x} = \phi(\theta,x)$ represents the output of the neural network, and $\phi(\theta,x):X \subseteq \mathbb{R}^{n} \to \mathcal{D} \subseteq \mathbb{R}^N$ is parameterized by $\theta$ with $n \ll N$. We map $x \in \mathbb{R}^n$ to a higher dimensional space $\mathbb{R}^N$ using $z$.
\end{definition}
\begin{definition}[Sphere Integral Function]
	\label{d3}
	The spherical integral, denoted by $\mathcal{B}$, represents the integration over a sphere with a radius of $\epsilon$ and a center point of $z_{\tilde{x}}$.
	\begin{equation}
		\mathcal{B}(z_{\tilde{x}},\epsilon)=\{z_{\tilde{x}}; \, \Vert z_x -z_{\tilde{x}} \Vert_2 \leq \epsilon\}.
	\end{equation}
\end{definition}
\begin{theorem}
	\label{t1}
	The expected risk function in a joint probability distribution can also be calculated as follows (Proof in Appendix \ref{app:proofs:t1}):
	\begin{equation}
		R(\phi) = 1- \mathbb{E}_{z_{\tilde{x}} \sim p(z_{\tilde{x}})}\int_{\mathcal{B}(z_{\tilde{x}},\epsilon)}p(z_x|z_{\tilde{x}})dz_x.
	\end{equation}
\end{theorem}
\begin{remark}
	The proof of Theorem \ref{t1} (in Appendix \ref{app:proofs:t1}) demonstrates that we can transform the problem of minimizing the expected risk function $R(\phi)$ into the problem of maximizing the probabilistic expectation $\mathbb{E}_{z_{\tilde{x}} \sim p(z_{\tilde{x}})}\int_{\mathcal{B}(z_{\tilde{x}},\epsilon)}p(z_x|z_{\tilde{x}})dz_x$ over a sphere integral domain $\mathcal{B}(z_{\tilde{x}},\epsilon)$. By finding an optimal value of $z_{\tilde{x}}$, denoted as $z_{\tilde{x}}^*$, that maximizes this probabilistic expectation, we effectively promote the minimization of $R(\phi)$. This optimization problem can be expressed as:
\end{remark}
\begin{equation}
	\label{e5}
	z_{\tilde{x}} = \argmax_{z_{\tilde{x}}^*\in \mathcal{D}_S}\int_{\mathcal{B}(z_{\tilde{x}}^*,\epsilon)}p(z_x|z_{\tilde{x}}^*)dz_x.
\end{equation}
\subsection{Bayesian Optimal Condensation Risk Function}
\label{sec:3.3}
\begin{theorem}
	\label{t2}
	The optimal synthetic image $z_{\tilde{x}}$ can be computed as follows (Proof in Appendix \ref{app:proofs:t2}):
	\begin{equation}
		\label{e6}
		z_{\tilde{x}} = \argmax_{z_{\tilde{x}}^*\in \mathcal{D}_S}\int_{\mathcal{B}(z_{\tilde{x}}^*,\epsilon)}\left[\log p(z_{\tilde{x}}^*|z_x)+\log p(z_x)\right]dz_x.
	\end{equation}
\end{theorem}
\begin{remark}
	By applying Bayes' rule and Jensen's inequality, we derive Eq. \eqref{e5}, which provides the formulaic representation for the log-likelihood and prior of the probability distribution as shown in Eq. \eqref{e6}. To obtain the realization of the random variable $D_S=z_{\tilde{x}}^*$, the objective is to find a series $z_x$ within the spherical region $\mathcal{B}$ that maximizes the integral function (Eq. \eqref{e6}). It is important to note that as $\epsilon$ approaches zero, $z_{\tilde{x}}$ approaches $\arg\max_{z_{x}}p(z_x|z_{\tilde{x}})$, which represents the solution for minimizing $R(\phi)$ by matching probability distributions. Since the solution for $p(z_x|z_{\tilde{x}}^*)$ cannot be obtained directly, the Bayesian formula is employed to rewrite it. To further investigate the lower bound of the function, we employ Jensen's inequality as an approximation method.
\end{remark}
\subsection{Approximating the Optimal Solution for Bayesian Condensation}
\label{sec:3.4}
The Eq. \eqref{e6} offers an optimal condensation solution within the Bayesian framework. However, its practical application faces three challenges. Firstly, computing the integral over the spherical region $\mathcal{B}(z_{\tilde{x}}, \epsilon)$ is challenging. Secondly, it is generally not possible to obtain the closed-form likelihood $ p(z_{\tilde{x}}^*|z_{x_i})$. Lastly, knowing the exact prior distribution $\log p(z_{x_i})$ is necessary.

\textbf{Monte Carlo Approximation: } To tackle the initial obstacle, we employ Monte Carlo sampling \cite{b68} to discretize the continuous expression by uniformly sampling $k$ points, represented as $z_{x_1},\dots,z_{x_k}$. The resultant discrete form of the expression is acquired through the subsequent procedure:
\begin{align}
	\label{e7}
	\frac{1}{k}\sum_{i=1}^{k} \log p(z_{\tilde{x}}^*|z_{x_i})+\log p(z_{x_i}).
\end{align} 

\begin{Assumption}[Likelihood Conforming Gaussian]
	\label{ass1}
	To estimate the log-likelihood $\log p(z_{\tilde{x}}^*|z_{x_i})$, we make the assumption that $p(z_{\tilde{x}}^*|z_{x_i})$ conforms to a Gaussian distribution. In this distribution, $\sigma_x^2$ represents the variance and $z_{x_i}$ represents the mean. It is denoted as $p(z_{\tilde{x}}^*|z_{x_i}) \sim \mathcal{N}(z_{x_i}, \sigma_{xi}^2 I)$.
\end{Assumption}
\begin{Assumption}[Prior Distribution Approximation with TV Extension]
	\label{ass2}
	The Total Variation (TV) and CLIP operation are incorporated as distribution priors to represent $\log p(z_{x_i})$, following the approach of Mislav {\it et al.} \cite{b49}. The CLIP operation constrains the probability within the bound of $[0,1]$. In contrast to their study, we extend the TV from a pixel-wise approach to a distribution-wise approach, which is also referred to as the total variation of probability distribution measures.
\end{Assumption}
Under the Assumption \ref{ass1} and Assumption \ref{ass2}, we divide Eq. \eqref{e7} into three separate loss terms as follows:
\begin{align}
	\label{e8}
	&\mathcal{L}_{\text{LH}} = -\frac{1}{2}\log(2 \pi \sigma_{x}) -\frac{1}{2\sigma_{x}^2}\Vert z_{\tilde{x}}^*- z_{x}\Vert_2^2,\\
	\label{e9}
	&\mathcal{L}_{\text{TV}} = \frac{1}{2} \Vert z_{\tilde{x}}^*- z_{x}\Vert_1, \\
	\label{e10}
	&\mathcal{L}_{\text{CLIP}} =  \left[\frac{z_{\tilde{x}}^*-z_{x}}{\sigma_{x}}-\text{CLIP}\left(\frac{z_{\tilde{x}}^*-z_{x}}{\sigma_{x}},0,1\right)\right]^2.
\end{align}
\subsection{Overall Loss Function and Pseudocode}
\begin{algorithm}[!ht]
	\caption{\underline{\textbf{BA}}yesian optimal \underline{\textbf{CON}}densation framework (\underline{\textbf{BACON}})}
	\label{alg1}
	\begin{algorithmic}[1]
	\REQUIRE Original dataset: $\mathcal{T}$, neural network: $\phi(\theta,x)$, probability distribution outputted by neural network: $z_x$ and $z_{\tilde{x}}$, where $x$ is the original image and $\tilde{x}$ is the synthetic image.
	\ENSURE Synthetic dataset $\mathcal{S}$.
	\STATE Initialize $x_1$ from $\mathcal{T}$ by randomly sampling.
	\FOR{$i = 1$ to $m-1$}
		\STATE Sample $x_1,\dots, x_k$ uniformly from $\mathcal{B}(z_{\tilde{x}}^*,\epsilon)$
		\STATE Calculate $\mathcal{L}_{\text{LH}}$ using Eq. \eqref{e8}
		\STATE Calculate $\mathcal{L}_{\text{TV}}$ according to Eq. \eqref{e9}
		\STATE Calculate $\mathcal{L}_{\text{CLIP}}$ by Eq. \eqref{e10} 
		\STATE Update the condensed dataset $\mathcal{S}$ by $\tilde{x}_{i+1}\gets \tilde{x}_{i}  + \alpha \nabla_{z_{\tilde{x}}}  \mathcal{L}_{\text{TOTAL}}$, where $\mathcal{L}_{\text{TOTAL}}$ is defined by Eq.~\eqref{e11}, and $\alpha$ denotes the learning rate for generating $\mathcal{S}$.
	\ENDFOR
	\STATE \textbf{return} $\mathcal{S}$
	\end{algorithmic}
\end{algorithm}
\label{sec:3.5}
\textbf{Overall Loss Function:} To summarize, the overall loss function of BACON integrates Eq. \eqref{e8}, Eq. \eqref{e9}, and Eq. \eqref{e10}. The expression for this combined loss function can be defined as follows:
\begin{equation}
	\label{e11}
	\mathcal{L}_{\text{TOTAL}} = \mathcal{L}_{\text{LH}} + \lambda \mathcal{L}_{\text{TV}} + (1-\lambda)\mathcal{L}_{\text{CLIP}},
\end{equation}
where the hyperparameter $\lambda$ serves as the weighting factor for the total loss function and is adjustable. By tuning $\lambda$, we can customize the loss function to optimize performance.

\textbf{Pseudocode Description:} The pseudocode of our algorithm is presented in Algorithm \ref{alg1}. Our proposed BACON is incorporated into the optimization pipeline to effectively guide algorithm optimization and enhance performance. The input to the algorithm comprises the original dataset $\mathcal{T}$, the neural network $\phi(\theta,x)$, and the probability distributions outputted by the neural network $z_x$ and $z_{\tilde{x}}$, where $x$ represents the original image and $\tilde{x}$ denotes the synthetic image. The output of the algorithm is the synthetic dataset $\mathcal{S}$. In the algorithm, the initial image $x_1$ is randomly sampled from the original dataset $\mathcal{T}$. Then, in each iteration, $k$ images $x_1,\dots,x_k$ are uniformly sampled from $\mathcal{B}(z_{\tilde{x}}^*,\epsilon)$ for each $i$. Subsequently, the total loss function $\mathcal{L}_{\text{TOTAL}}$ is calculated based on three loss functions: $\mathcal{L}_{\text{LH}}$, $\mathcal{L}_{\text{TV}}$, and $\mathcal{L}_{\text{CLIP}}$. Finally, the synthetic dataset $\mathcal{S}$ is updated by generating the next synthetic image $\tilde{x}_{i+1}$.
\section{Experimental Evaluation}
\label{sec:experiment}
To demonstrate the effectiveness of BACON, we conducted extensive experiments on both large-scale and small-scale benchmark datasets. Furthermore, comprehensive comparative experiments and ablation studies were performed to further assess its performance. Finally, we presented the outcomes of BACON through visualization. Additional information about our experiments can be found in Appendix \ref{app:experiment}.
\subsection{Experiment Setup}
For a systematic evaluation of our method, we assess its efficacy through experiments conducted on widely-used dataset distillation benchmarks, including the MNIST \cite{b58}, Fashion-MNIST \cite{b60}, SVHN \cite{b61}, CIFAR-10/100 \cite{b52}, and TinyImageNet \cite{b53}. We employ the ConvNet architecture \cite{b54} for dataset distillation experiments, following prior research approaches \cite{b11, b34}. The performance of the synthetic dataset is assessed by averaging the top-1 accuracy of the trained model over five experiments on the validation set with IPC-50, IPC-10, and IPC-1, respectively. We set $\lambda$ in Eq. \eqref{e11} to $0.8$ in our experiments, except for ablation studies. Additionally, we maintain consistency with IDM \cite{b56} for most hyperparameter settings, and the basic environment adheres to the guidelines outlined in DC-bench \cite{b55}. Further implementation details can be found in Appendix \ref{app:experiment:setup}.
\subsection{Comparison to the State-of-the-art Methods}
\begin{figure*}
	\centering
	\includegraphics[width=0.95\linewidth]{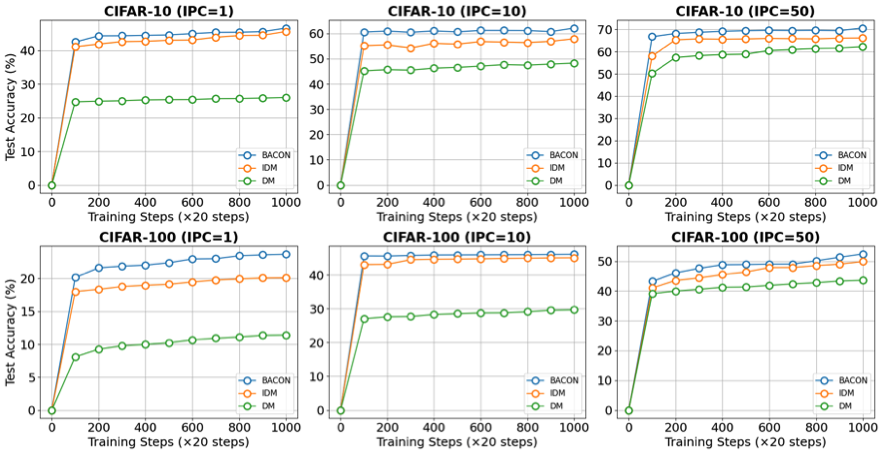}
	\caption{\textbf{Performance comparison with BACON, IDM, and DM across varying training steps on the CIFAR-10/100 datasets:} The blue line with white circles represents our proposed BACON, the orange line with white circles represents IDM, and the green line with white circles represents DM. All synthetic images are generated using the CIFAR-10/100 datasets across training steps from 0 to 20000 with IPC-1, IPC-10, and IPC-50, respectively. }
	\label{fig3}
  \end{figure*}
\renewcommand{\arraystretch}{1} 
\setlength{\tabcolsep}{2pt} 
\setlength{\extrarowheight}{1pt} 
    \begin{table*}[htb]
        \centering
        \Huge
    \caption{\textbf{Comparison with previous coreset selection and dataset condensation methods:} Like most state-of-the-art methods, we evaluate our method on six datasets (MNIST, Fashion-MNIST, SVHN, CIFAR-10/100, TinyImageNet) with different numbers of synthetic images per class (IPC). The ``Ratio(\%)'' represents the condensed images' ratio to the entire training set. For reference, ``Full Set'' indicates the accuracy of the trained model on the complete training set. It's important to note that DD and LD employ different architectures, specifically LeNet \cite{b58} for MNIST and AlexNet \cite{b2} for CIFAR-10. Meanwhile, the remaining methods all utilize ConvNet \cite{b54}.}
        \resizebox{\textwidth}{!}{
        \begin{tabular}{c|ccc|cccccc|ccc|ccc|cc}
        \toprule
                   & \multicolumn{3}{c|}{MNIST}       & \multicolumn{3}{c|}{Fashion-MNIST}        & \multicolumn{3}{c|}{SVHN}       & \multicolumn{3}{c|}{CIFAR-10}            & \multicolumn{3}{c|}{CIFAR-100}            & \multicolumn{2}{c}{TinyImageNet} \\ \midrule
        IPC        & 1           & 10         & 50    & 1     & 10    & \multicolumn{1}{c|}{50}   & 1     & 10          & 50        & 1           & 10          & 50          & 1           & 10          & 50           & 1               & 10             \\
        Ratio (\%) & 0.017       & 0.17       & 0.83  & 0.017 & 0.17  & \multicolumn{1}{c|}{0.83} & 0.014 & 0.14        & 0.68      & 0.02        & 0.2         & 1           & 0.2         & 2           & 10           & 0.2             & 2              \\ \midrule
        Random \cite{b59}    & 64.9        & 95.1       & 97.9  & 51.4  & 73.8  & \multicolumn{1}{c|}{82.5} & 14.6  & 35.1        & 70.9      & 14.4        & 26          & 43.4        & 4.2         & 14.6        & 30           & 1.4             & 5              \\
        Herding \cite{b41}   & 89.2        & 93.7       & 94.8  & 67    & 71.1  & \multicolumn{1}{c|}{71.9} & 20.9  & 50.5        & 72.6      & 21.5        & 31.6        & 40.4        & 8.4         & 17.3        & 33.7         & 2.8             & 6.3            \\ \midrule
        DD \cite{b22}        & -          & 79.5       & -     & -     & -     & \multicolumn{1}{c|}{-}    & -     & -           & -         & -           & 36.8        & -           & -           & -           & -            & -               & -              \\
        LD \cite{b57}        & 60.9        & 87.3       & 93.3  & -     & -     & \multicolumn{1}{c|}{-}    & -     & -           & -         & 25.7        & 38.3        & 42.5        & 11.5        & -           & -            & -               & -              \\ \midrule
        DC \cite{b28}        & 91.7        & 97.4       & 98.8  & 70.5  & 82.3  & \multicolumn{1}{c|}{83.6} & 31.2  & 76.1        & 82.3      & 28.3        & 44.9        & 53.9        & 12.8        & 26.6        & 32.1         & 5.3             & 11.1           \\
        DSA \cite{b33}       & 88.7        & 97.8       & 99.2  & 70.6  & 86.6  & \multicolumn{1}{c|}{88.7} & 27.5  & 79.2        & 84.4      & 28.8        & 53.2        & 60.6        & 13.9        & 32.3        & 42.8         & 6.6             & 16.3           \\
        DCC \cite{b62}       & -           & -          & -     & -     & -     & \multicolumn{1}{c|}{-}    & 47.5  & 80.5        & 80.5      & 34          & 54.5        & 64.2        & 14.6        & 33.5        & 39.3         & -               & -              \\
        CAFE(DSA) \cite{b32} & 90.8        & 97.5       & 98.9  & 73.7  & 83    & \multicolumn{1}{c|}{88.2} & 42.9  & 77.9        & 82.3      & 31.6        & 50.9        & 62.3        & 14          & 31.5        & 42.9         & -               & -              \\
        DM \cite{b11}        & 89.2        & 97.3       & 94.8  & -     & -     & \multicolumn{1}{c|}{-}    & -     & -           & -         & 26          & 48.9        & 63          & 11.4        & 29.7        & 43.6         & 3.9             & 12.9           \\ \midrule
        \rowcolor[HTML]{FFFFE0}
        IDM \cite{b56}       & 93.82       & 96.26      & 97.01    & 78.23    & 82.53    & \multicolumn{1}{c|}{84.03}   & \textbf{69.45}    & 82.95       & 87.5      & 45.60        & 58.6        & 67.5        & 20.1        & 45.1        & 50           & 10.1            & 21.9          \\
        \rowcolor[HTML]{FFFFE0}
        BACON [Ours]     & \begin{tabular}[c]{@{}c@{}}\textbf{94.15}\\ (\red{0.33 $\uparrow$})\end{tabular} & \begin{tabular}[c]{@{}c@{}}\textbf{97.3}\\ (\red{1.04 $\uparrow$})\end{tabular} & \begin{tabular}[c]{@{}c@{}}\textbf{98.01}\\ (\red{1.00 $\uparrow$})\end{tabular} & \begin{tabular}[c]{@{}c@{}}\textbf{78.48}\\ (\red{0.25 $\uparrow$})\end{tabular} & \begin{tabular}[c]{@{}c@{}}\textbf{84.23}\\ (\red{1.70 $\uparrow$})\end{tabular} & \multicolumn{1}{c|}{\begin{tabular}[c]{@{}c@{}}\textbf{85.52}\\ (\red{1.49 $\uparrow$})\end{tabular}}   & \begin{tabular}[c]{@{}c@{}}69.44\\ (\green{0.01 $\downarrow$})\end{tabular} & \begin{tabular}[c]{@{}c@{}}\textbf{84.64}\\ (\red{1.69 $\uparrow$})\end{tabular} & \begin{tabular}[c]{@{}c@{}}\textbf{89.1}\\ (\red{1.60 $\uparrow$})\end{tabular} & \begin{tabular}[c]{@{}c@{}}\textbf{45.62}\\ (\red{0.02 $\uparrow$})\end{tabular} & \begin{tabular}[c]{@{}c@{}}\textbf{62.06}\\ (\red{3.46 $\uparrow$})\end{tabular} & \begin{tabular}[c]{@{}c@{}}\textbf{70.06}\\ (\red{2.56 $\uparrow$})\end{tabular} & \begin{tabular}[c]{@{}c@{}}\textbf{23.68}\\ (\red{3.58 $\uparrow$})\end{tabular} & \begin{tabular}[c]{@{}c@{}}\textbf{46.15}\\ (\red{1.05 $\uparrow$})\end{tabular} & \begin{tabular}[c]{@{}c@{}}\textbf{52.29}\\ (\red{2.29 $\uparrow$})\end{tabular} & \begin{tabular}[c]{@{}c@{}}\textbf{10.2}\\ (\red{0.1 $\uparrow$})\end{tabular}      & \begin{tabular}[c]{@{}c@{}}\textbf{25}\\ (\red{3.1 $\uparrow$})\end{tabular}     \\ \midrule
        Full Set   & \multicolumn{3}{c|}{99.6}        & \multicolumn{3}{c|}{93.5}                  & \multicolumn{3}{c|}{95.4}       & \multicolumn{3}{c|}{84.8}               & \multicolumn{3}{c|}{56.2}                & \multicolumn{2}{c}{37.6}         \\ \bottomrule
        \end{tabular}
        }
        \label{tab1}
    \end{table*}

For the convenience of analyzing the performance of our method across multiple datasets, we categorized the datasets into three groups based on their resolution: (1) low resolution (MNIST and F-MNIST); (2) medium resolution (SVHN, CIFAR-10, and CIFAR-100); and (3) high resolution (TinyImageNet). To assess its performance, we compared BACON with state-of-the-art methods such as core-set selection and dataset distillation methods. The results are illustrated in Table \ref{tab1}, and the performance comparison of BACON, IDM, and DM across varying training steps on CIFAR-10/100 datasets is shown in Figure \ref{fig3}. For more details on comparable experiments, please refer to Appendix \ref{app:experiment:comparison}.
\paragraph{Analysis}
\begin{table}[h]
    \begin{minipage}{0.5\textwidth}
    \centering
    \caption{\textbf{Ablation study of diverse loss functions:} Evaluating the performance of the proposed loss functions $\mathcal{L}_\text{LH}$, $\mathcal{L}_\text{TV}$, and $\mathcal{L}_\text{CLIP}$ individually. All experimental hyperparameters, represented by $\lambda$, are set to 0.8 by default. This experiment focuses on the CIFAR-10 dataset with an IPC setting of 50.}
    \begin{tabular}{cccc}
    \toprule
    $\mathcal{L}_\text{LH}$ & $\mathcal{L}_\text{TV}$ & $\mathcal{L}_\text{CLIP}$ & Test acc. (\%) \\ 
    \midrule
    \cmark & \xmark& \xmark& 64.86 \\
    \rowcolor[HTML]{D3D3D3}
    \xmark& \cmark & \xmark& 69.96 \\
    \xmark& \xmark& \cmark & 55.07 \\
    \midrule
    \rowcolor[HTML]{D3D3D3}
    \cmark & \cmark & \xmark& 69.81 \\
    \cmark &\xmark & \cmark & 64.78 \\
    \xmark& \cmark & \cmark & 69.76 \\
    \midrule
    \rowcolor[HTML]{D3D3D3}
    \rcmark &\rcmark & \rcmark & \red{\textbf{70.06}} \\
    \bottomrule
    \end{tabular}
    \label{tab2}
    \end{minipage}%
    \begin{minipage}{0.5\textwidth}
        \begin{figure}[H]
            \centering
            \includegraphics[width=0.8\linewidth]{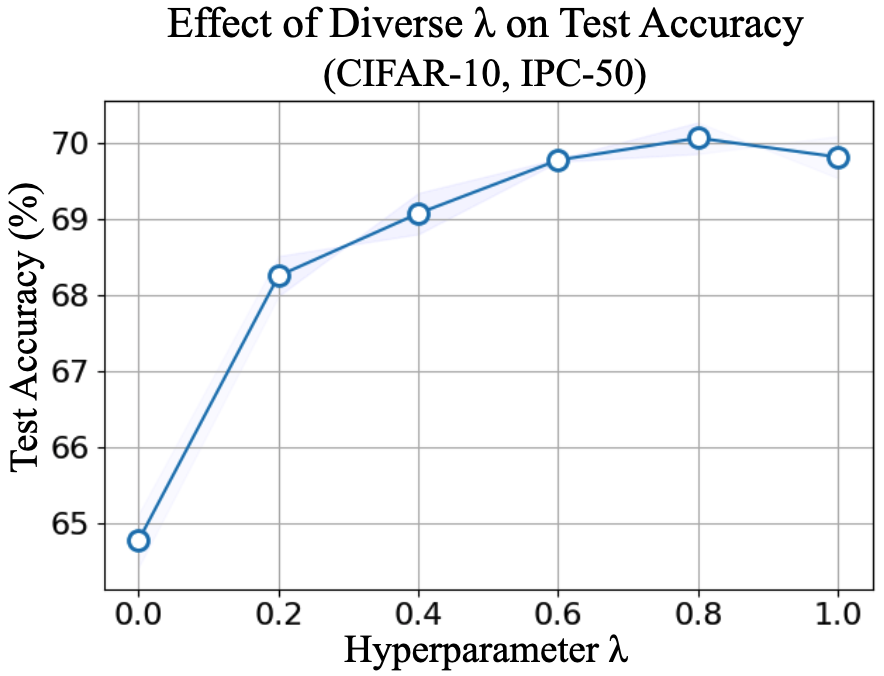}
            \caption{\textbf{Ablation study of diverse hyperparameters: } Sampling diverse hyperparameters from $\lambda = [0.0,1.0]$ and obtaining the effectiveness of diverse $\lambda$ on the test accuracy with the CIFAR-10 dataset and 50 images per class (IPC-50).}
            \label{fig4}
        \end{figure}
    \end{minipage}
    \begin{minipage}{1\textwidth}
        \begin{figure}[H]
            \centering
            \includegraphics[width=0.8\linewidth]{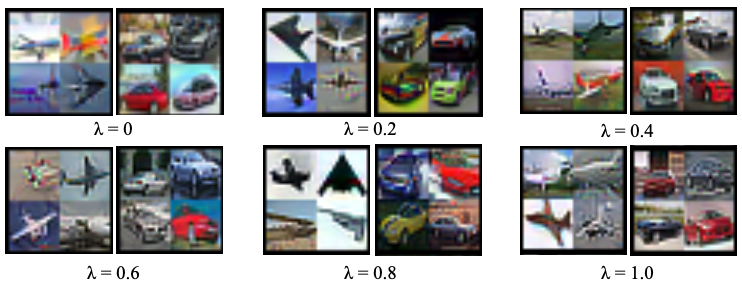}
            \caption{\textbf{Visualization of diverse hyperparameters: } Visualizing synthetic images generated with diverse hyperparameters on CIFAR-10 test accuracy, using 50 images per class (IPC-50). The left image of a pair represents an airplane, and the right one represents an automobile.}
            \label{fig5}
        \end{figure}
    \end{minipage}
\end{table}
On low-resolution datasets, BACON outperforms IDM across all IPC settings. Specifically, with IPC-1, BACON achieves an accuracy of 94.15\% on the MNIST dataset and 78.48\% on the F-MNIST dataset, which are the highest accuracies compared to other methods. The performance improvement is comparatively limited under the IPC settings of 10 or 50, primarily due to the constraints imposed by our method built upon IDM. On medium-resolution datasets, BACON achieves the highest accuracy of 89.1\% on SVHN and 70.06\% on CIFAR-10, which are close to the performance of models trained on the original datasets. Our method generally outperforms others on medium-resolution datasets, with negligible performance degradation observed for IPC-1 on SVHN, decreasing by only 0.01. The most likely reason for this is that low resolution and a low IPC setting can cause the synthetic dataset to deviate from its Gaussian distribution, undermining the accuracy of our approximate solution in accurately describing the optimization direction. This deviation is evident in SVHN with IPC-1 and CIFAR-10 with IPC-1. Ultimately, BACON slightly outperforms other methods with IPC-1 in high-resolution scenarios and significantly surpasses them with a higher IPC setting. This finding further supports our inference that the approximate solution space of BACON may lead to optimization in the wrong direction under low-resolution and low IPC settings.
\subsection{Ablation Studies}
\paragraph{Effectiveness of Diverse Loss Functions}
The impacts of the three loss function terms in BACON, namely $\mathcal{L}_\text{LH}$, $\mathcal{L}_\text{TV}$, and $\mathcal{L}_\text{CLIP}$, on test accuracy are presented in Table \ref{tab2}. Synthetic datasets were generated on CIFAR-10 using diverse loss terms with an IPC-50 setting by BACON. IDM and DM achieved test accuracies of 67.5\% and 63\%, respectively, under similar experimental settings. However, our proposed BACON achieved the highest accuracy of 70.06\% with $\mathcal{L}_\text{TOTAL}$. When utilizing only one loss term from $\mathcal{L}_\text{LH}$, $\mathcal{L}_\text{TV}$, and $\mathcal{L}_\text{CLIP}$, $\mathcal{L}_\text{TV}$ resulted in the highest accuracy (69.96\%), followed by $\mathcal{L}_\text{LH}$ (64.86\%), and $\mathcal{L}_\text{CLIP}$ (55.07\%). When two loss terms are used from $\mathcal{L}_\text{LH}$, $\mathcal{L}_\text{TV}$, and $\mathcal{L}_\text{CLIP}$, $\mathcal{L}_\text{LH}+\mathcal{L}_\text{TV}$ resulted in the highest accuracy (69.81\%), followed by $\mathcal{L}_\text{TV}+\mathcal{L}_\text{CLIP}$ (69.76\%), and $\mathcal{L}_\text{LH}+\mathcal{L}_\text{CLIP}$ (64.78\%). From the data above, it is evident that $\mathcal{L}_\text{TV}$ significantly contributes to improving accuracy, while the other two terms do not contribute as much. Remarkably, amalgamating all three terms results in the highest performance, as evidenced by the findings presented in the final row of Table \ref{tab2}. Further details regarding the ablation studies can be found in the Appendix \ref{app:experiment:ablation}.
\paragraph{Effectiveness of Diverse Hyperparameter $\lambda$}
We explored the impact of diverse hyperparameters $\lambda$ ranging from 0 to 1 on the CIFAR-10 dataset using the IPC-50 setting with BACON. The performance of the synthetic dataset under different $\lambda$ values is illustrated in Figure \ref{fig4}. We observed a steady increase in test accuracy with increasing values of $\lambda$ until it peaked at $\lambda = 0.8$, where the accuracy exceeded 70\%. Beyond this point, further increments in $\lambda$ did not achieve significant improvements in accuracy. Therefore, we infer that the loss function exhibits its highest efficacy when $\lambda$ is set to 0.8. The visualization of various hyperparameters is depicted in Figure \ref{fig5}.
\subsection{Visualization}
\begin{figure*}
	\centering
	\includegraphics[width=0.84\linewidth]{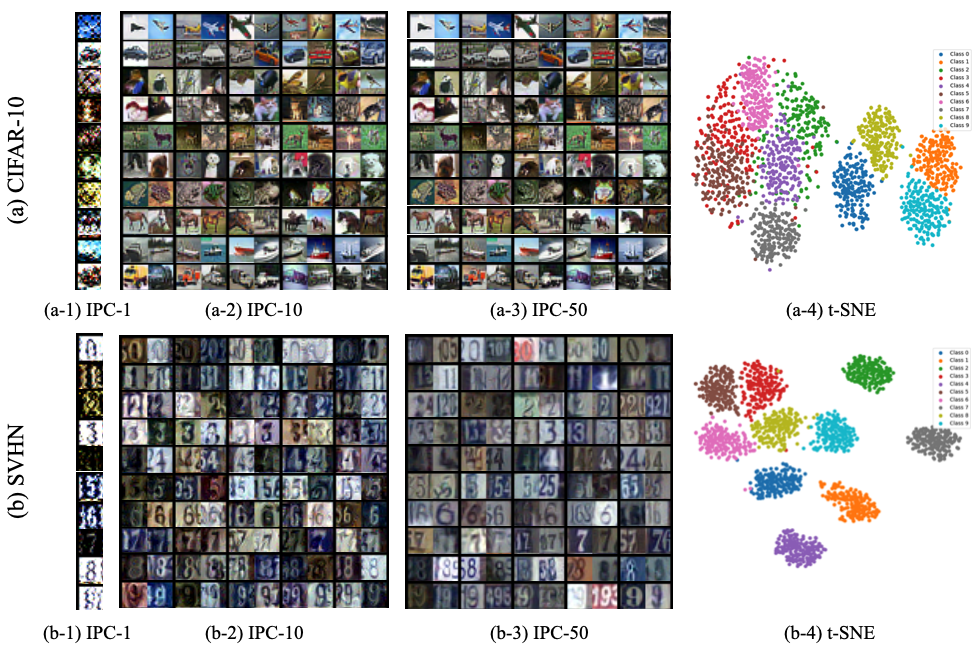}
	\caption{\textbf{Visualizations of BACON with IPC-1/10/50 on CIFAR-10 and SVHN datasets:} (a-1), (a-2), and (a-3) represent the synthetic images generated on the CIFAR-10 dataset with the settings of IPC-1, IPC-10, and IPC-50, respectively. (b-1), (b-2), and (b-3) represent the synthetic images generated on the SVHN dataset with the settings of IPC-1, IPC-10, and IPC-50. (a-4) and (b-4) denote the visualizations of clusters of classes by t-SNE.}
	\label{fig6}
  \end{figure*}
  The outcomes of BACON are visualized in Figure \ref{fig6}. Specifically, (a-1), (a-2), and (a-3) represent synthetic samples with IPC-1, IPC-10, and IPC-50, respectively, on CIFAR-10, while (b-1), (b-2), and (b-3) represent samples on SVHN. Additionally, Figure \ref{fig6} (a-4) and Figure \ref{fig6} (b-4) display class cluster visualizations for these two datasets using t-SNE. As IPC numbers increase, so does the compression rate, leading to richer visual information in the images. This is because synthetic examples are not limited to real examples. IPC-1 generates more representative images by aggregating semantic information from all original training sets, while IPC-50 images closely resemble the originals, with IPC-10 falling in between. SVHN demonstrates better clustering than CIFAR-10. Further experiment visualization details, including comparisons with previous methods, can be found in Appendix \ref{app:experiment:visualization}. 
\section{Conclusion}
\label{sec:conclusion}
In this paper, we addressed two key challenges in Dataset Distillation (DD): 1) how to effectively formulate the DD problem, and 2) determining the theoretical lower bound of optimal condensation. To tackle these issues, we introduced BACON, a novel and efficient approach leveraging Bayesian principles. BACON provides the first formal analysis of the DD problem, yielding both an optimal solution and its approximate form. Extensive experiments across multiple datasets demonstrate that BACON consistently outperforms existing state-of-the-art methods. Compared to the IDM method, BACON achieves its highest accuracy gains as follows: 1.70\% on low-resolution datasets, 3.58\% on medium-resolution datasets, and 3.10\% on high-resolution datasets. Ablation studies further confirm the effectiveness of BACON. These results underscore the practical applicability and solid theoretical foundation of our approach, paving the way for future research. Our study opens new avenues for extending BACON to more complex tasks and refining its theoretical framework.

\newpage

\bibliographystyle{plainnat}
\bibliography{main}

\newpage
\appendix
\begin{center}
	{\Large \textbf{Supplementary Material\\}} \vskip 0.1in \textbf{BACON: Bayesian Optimal Condensation Framework \\for Dataset Distillation}
  \end{center}
  \vskip 0.3in
  \begin{itemize}
    \item Appendix \ref{app:proofs} contains proofs for all theorems and assumptions presented in this paper.
    \item Appendix \ref{app:extended background} provides additional background information and preliminary details on dataset distillation. 
    \item Appendix \ref{app:experiment} provides implementation details of experiments and visualizations.
    \item Appendix \ref{app:impact} outlines the potential social impact of our work;
    \item Appendix \ref{app:limitation} explores the limitations of our work and outlines potential future directions.
  \end{itemize}
\section{Proofs}
\label{app:proofs}
\subsection{Proof of Theorem \ref{t1}}
\label{app:proofs:t1}
\begin{definition}[Similarity Indicator of $\epsilon$-neighborhood]
	\label{proof_d1}
	Let $z_x$ and $z_{\tilde{x}}$ be two distributions. A binary loss function $\mathcal{L}$ is used to assess their similarity. When the Euclidean distance between the two distributions is small, $\mathcal{L}$ is equal to $0$; otherwise, $\mathcal{L}$ is equal to $1$. As $\epsilon$ approaches zero, the synthetic distribution $z_{\tilde{x}}$ approximates the original distribution $z_x$. We define the similarity indicator function as follows:
	\begin{equation}
		\mathcal{L}(z_x,z_{\tilde{x}}) := \mathbbm{1}\{\Vert z_x - z_{\tilde{x}} \Vert_2 \geq \epsilon\}.
	\end{equation}
\end{definition}
\begin{definition}[Expected Risk Function]
	\label{proof_d2}
	Consider a joint probability distribution $p(z_{x},z_{\tilde{x}})$ formed by the probability distribution of the outputs of neural networks, $\mathcal{D}_T$ and $\mathcal{D}_S$, for the original dataset $\mathcal{T}$ and the synthetic dataset $\mathcal{S}$. Given $z_{x} \in \mathcal{D}_T$ and $z_{{\tilde{x}}}\in \mathcal{D}_S$, the expected risk function in the joint probability distribution $R(\phi)$ is defined as follows:
	\begin{equation}
		R(\phi) := \mathbb{E}_{(z_{x},z_{\tilde{x}}) \sim p(z_{x},z_{\tilde{x}})}[\mathcal{L}(z_{x},z_{ \tilde{x}})],
	\end{equation}
	where $z_{x} = \phi(\theta,x)$ represents the output of the neural network, and $\phi(\theta,x):X \subseteq \mathbb{R}^{n} \to \mathcal{D} \subseteq \mathbb{R}^N$ is parameterized by $\theta$ with $n \ll N$. We map $x \in \mathbb{R}^n$ to a higher dimensional space $\mathbb{R}^N$ using $z$.
\end{definition}
\begin{definition}[Sphere Integral Function]
	\label{proof_d3}
	The spherical integral, denoted by $\mathcal{B}$, represents the integration over a sphere with a radius of $\epsilon$ and a center point of $z_{\tilde{x}}$.
	\begin{equation}
		\mathcal{B}(z_{\tilde{x}},\epsilon)=\{z_{\tilde{x}}; \, \Vert z_x -z_{\tilde{x}} \Vert_2 \leq \epsilon\}.
	\end{equation}
\end{definition}
\begin{theorem}
	The expected risk function in a joint probability distribution can also be calculated as follows:
	\begin{equation}
		R(\phi) = 1- \mathbb{E}_{z_{\tilde{x}} \sim p(z_{\tilde{x}})}\int_{\mathcal{B}(z_{\tilde{x}},\epsilon)}p_(z_x|z_{\tilde{x}})dz_x.
	\end{equation}
	\begin{proof}\let\qed\relax Under the Definition \ref{d1},\ref{d2} and \ref{d3} (Definition \ref{proof_d1}, \ref{proof_d2} and \ref{proof_d3} in Appendix \ref{app:proofs:t1}), we derive the risk function in a joint probability distribution as follows:
		\begin{align}
			R(\phi) &= \mathbb{E}_{(z_x,z_{\tilde{x}}) \sim p(z_x,z_{\tilde{x}})}[\mathcal{L}(z_x,z_{\tilde{x}})]\\
			&=\mathbb{E}_{z_{\tilde{x}} \sim p(z_{\tilde{x}})}\mathbb{E}_{z_x \sim p_(z_x| z_{\tilde{x}})}[ \mathbbm{1}\{\Vert z_x - z_{\tilde{x}} \Vert_2 \geq \epsilon \}]\\
			&= \int_{\mathcal{D}_S}p(z_{\tilde{x}})\int_{\mathcal{D}_T} p(z_x|z_{\tilde{x}})\cdot \mathbbm{1}\{\Vert z_x - z_{\tilde{x}} \Vert_2 \geq \epsilon\}dz_xdz_{\tilde{x}}\\
			&= \int_{\mathcal{D}_S}p(z_{\tilde{x}})\int_{\mathcal{D}_T \setminus\mathcal{B}(z_{\tilde{x}},\epsilon)}p(z_x|z_{\tilde{x}})dz_xdz_{\tilde{x}}\\
			&= \int_{\mathcal{D}_S}p(z_{\tilde{x}})\left[1- \int_{\mathcal{B}(z_{\tilde{x}},\epsilon)}p(z_x|z_{\tilde{x}})dz_x\right] dz_{\tilde{x}}\\
			&= \int_{\mathcal{D}_S}p(z_{\tilde{x}})dz_{\tilde{x}}- \int_{\mathcal{D}_S}p(z_{\tilde{x}})\int_{\mathcal{B}(z_{\tilde{x}},\epsilon)}p(z_x|z_{\tilde{x}})dz_x dz_{\tilde{x}}\\
			&= 1- \int_{\mathcal{D}_S}p(z_{\tilde{x}})\int_{\mathcal{B}(z_{\tilde{x}},\epsilon)}p_(z_x|z_{\tilde{x}})dz_xdz_{\tilde{x}}\\
			&= 1- \mathbb{E}_{z_{\tilde{x}} \sim p(z_{\tilde{x}})}\int_{\mathcal{B}(z_{\tilde{x}},\epsilon)}p_(z_x|z_{\tilde{x}})dz_x.
		\end{align}
	\end{proof}
\end{theorem}
\begin{remark}
	The proof demonstrates that we can transform the problem of minimizing the expected risk function $R(\phi)$ into the problem of maximizing the probabilistic expectation $\mathbb{E}_{z_{\tilde{x}} \sim p(z_{\tilde{x}})}\int_{\mathcal{B}(z_{\tilde{x}},\epsilon)}p(z_x|z_{\tilde{x}})dz_x$ over a sphere integral domain $\mathcal{B}(z_{\tilde{x}},\epsilon)$. By finding an optimal value of $z_{\tilde{x}}$, denoted as $z_{\tilde{x}}^*$, that maximizes this probabilistic expectation, we effectively promote the minimization of $R(\phi)$. This optimization problem can be expressed as:
\end{remark}
\subsection{Proof of Theorem \ref{t2}}
\label{app:proofs:t2}
\textbf{Bayes Rule: } The Bayes' rule can be defined as:
\begin{equation}
	P(A|B) = \frac{P(B|A)P(A)}{P(B)},
\end{equation}
where $A$ and $B$ are events and $P(B) \neq 0 $.\\
\textbf{Jensen's Inequality: } The Jensen's inequality can be written as:
\begin{equation}
	\varphi(\mathbb{E}[X]) \leq \mathbb{E}[\varphi(X)],
\end{equation}
where $\varphi$ is a linear function defined on a convex set. If $\varphi$ is a linear function defined on a concave set, Jensen's inequality can be defined as follows:
\begin{equation}
	\varphi(\mathbb{E}[X]) \geq \mathbb{E}[\varphi(X)].
\end{equation}
\textbf{Proof of Convexity of Logarithmic Function: } The function is concave if its second derivative is negative. We have that
\begin{equation}
	\frac{\partial^2g(x)}{\partial^2x} = \frac{\partial}{\partial x}\left(\frac{\partial g(x)}{\partial x}\right) = \frac{\partial}{\partial x}\left( \frac{\partial \log(x)}{\partial x}\right) = \frac{\partial}{\partial x}\left( \frac{1}{x} \right) = - \frac{1}{x^2}.
\end{equation}
For $x > 0$. Hence, $g(x)$ is a concave function.
\begin{theorem}
	The optimal synthetic image $z_{\tilde{x}}$ can be computed as follows:
	\begin{equation}
		z_{\tilde{x}} = \argmax_{z_{\tilde{x}}^*\in \mathcal{D}_S}\int_{\mathcal{B}(z_{\tilde{x}}^*,\epsilon)}\left[\log p(z_{\tilde{x}}^*|z_x)+\log p(z_x)\right]dz_x.
	\end{equation}
	\begin{proof}\let\qed\relax By leveraging the Bayes rule and Jensen's inequality, we derive the function as follows: 
		\begin{align}
			\label{e29}
			z_{\tilde{x}} &= \argmax_{z_{\tilde{x}}^*\in \mathcal{D}_S}\int_{\mathcal{B}(z_{\tilde{x}}^*,\epsilon)}p(z_x|z_{\tilde{x}}^*)dz_x\\
			&= \argmax_{z_{\tilde{x}}^*\in \mathcal{D}_S}\int_{\mathcal{B}(z_{\tilde{x}}^*,\epsilon)}\frac{p(z_{\tilde{x}}^*|z_x) p(z_x)}{p(z_{\tilde{x}}^*)}dz_x\\
			&=\argmax_{z_{\tilde{x}}^*\in \mathcal{D}_S}\int_{\mathcal{B}(z_{\tilde{x}}^*,\epsilon)}\underbrace{ p(z_{\tilde{x}}^*|z_x) p(z_x)}_{\text{Bayesian formula}}dz_x\\
			&= \argmax_{z_{\tilde{x}}^*\in \mathcal{D}_S}\left[\log \int_{\mathcal{B}(z_{\tilde{x}}^*,\epsilon)}p(z_{\tilde{x}}^*|z_x) p(z_x)dz_x\right]\\
			&\underbrace{\geq \argmax_{z_{\tilde{x}}^*\in \mathcal{D}_S}\int_{\mathcal{B}(z_{\tilde{x}}^*,\epsilon)}\log \left[p(z_{\tilde{x}}^*|z_x) p(z_x)\right]dz_x}_{\text{Jensen's inequality}}\\
			\label{e34}
			&= \argmax_{z_{\tilde{x}}^*\in \mathcal{D}_S}\int_{\mathcal{B}(z_{\tilde{x}}^*,\epsilon)}\left[\log p(z_{\tilde{x}}^*|z_x)+\log p(z_x)\right]dz_x.
		\end{align}
	\end{proof}
\end{theorem}
\begin{remark}
	By applying Bayes' rule and Jensen's inequality, we derive Eq. \eqref{e29}, which provides the formulaic representation for the log-likelihood and prior of the probability distribution as shown in Eq. \eqref{e34}. To obtain the realization of the random variable $D_S=z_{\tilde{x}}^*$, the objective is to find a series $z_x$ within the spherical region $\mathcal{B}$ that maximizes the integral function (Eq. \eqref{e34}). It is important to note that as $\epsilon$ approaches zero, $z_{\tilde{x}}$ approaches $\arg\max_{z_{x}}p(z_x|z_{\tilde{x}})$, which represents the solution for minimizing $R(\phi)$ by matching probability distributions. Since the solution for $p(z_x|z_{\tilde{x}}^*)$ cannot be obtained directly, the Bayesian formula is employed to rewrite it. To further investigate the lower bound of the function, we employ Jensen's inequality as an approximation method.
\end{remark}
\paragraph{Discussion}
The insights provided by Theorems \ref{t1} and \ref{t2} offer valuable contributions to the fields of synthetic image generation and probability distribution matching for dataset distillation. The approach outlined in Theorem \ref{t1}, which emphasizes maximizing probabilistic expectations within a spherical integral domain, presents a promising avenue for minimizing loss by aligning probability distributions. However, it's essential to acknowledge the potential limitations imposed by assumptions regarding distributions and integral domains in practical applications. On the other hand, Theorem \ref{t2} leverages foundational mathematical principles such as Bayes' rule and Jensen's inequality to establish a clear framework for determining the optimal synthetic image $z_{\tilde{x}}$. While this theorem provides valuable guidance, its applicability hinges on the validity of the underlying assumptions inherent in Bayes' rule and Jensen's inequality. In essence, these theorems significantly advance our understanding of image synthesis and probabilistic modeling. However, their real-world utility necessitates further validation and refinement through empirical experiments and practical applications.

\section{Extended Background}
\label{app:extended background}
\subsection{Dataset Distillation}
Let us denote the real dataset $\mathcal{T} = \{(x_i,y_i)\}_{i=1}^{|\mathcal{T}|}$, consisting of $|\mathcal{T}|$ pairs of training images and corresponding labels, where $x \in \mathcal{X}$ and $\mathcal{X} \subset \mathbb{R}^{d}$, $y \in \mathcal{Y}$ and $\mathcal{Y} = \{0, \dots, C-1\}$. $d$ is the number of features and $C$ is the number of classes. The synthetic dataset is denoted as $\mathcal{S} = \{(\tilde{x}_i,\tilde{y}_i)\}\vert ^{\vert\mathcal{S}\vert}_{i=1}$, where $\tilde{x} \in \mathbb{R}^{d}$, $\tilde{y} \in \mathcal{Y}$, and $\mathcal{S} \ll \mathcal{T}$.

Our objective is to map the original dataset $\mathcal{T}$ to the dataset $\mathcal{S}$, which is of lower magnitude, while still preserving the informative content. We aim to achieve this by using a differentiable function $\phi_\theta : x\rightarrow y$, where $\theta$ represents the parameters. The problem of DD can be formulated as follows:
\begin{equation}
	\mathbb{E}_{x\sim P_{\mathcal{D}}}[l(\phi_{\theta^{\mathcal{T}}}(x),y)]\simeq \mathbb{E}_{x\sim P_{\mathcal{D}}}[l(\phi_{\theta^{\mathcal{S}}}(\tilde{x}),\tilde{y})],
\end{equation}
where $x\sim P_{\mathcal{D}}$ represents the original dataset $x$ sampled from the test dataset $\mathcal{D} \subset \mathcal{T}$ and $l$ indicates the loss function, specifically the cross-entropy loss. The deep neural network is denoted as $\phi$, which is parameterized by $\theta$. Meanwhile, $\phi_{\theta^{\mathcal{T}}}$ and $\phi_{\theta^{\mathcal{S}}}$ refer to the networks that are trained on $\mathcal{T}$ and $\mathcal{S}$, respectively.
\subsection{Meta-learning Based Method}
Previous studies have mainly focused on treating the DD task as a meta-learning problem \cite{b22,b36}. In these studies, the network parameters $\theta^{\mathcal{S}}$ were represented as a function of the synthetic dataset $\mathcal{S}$. The solution for $\mathcal{S}$ was obtained by minimizing the training loss $\mathcal{L}^{\mathcal{T}}$ on the original dataset $\mathcal{T}$. The formulation can be provided as follows:
\begin{align}
	\mathcal{S}^* = &\arg\min_\mathcal{S} \mathcal{L}^{\mathcal{T}}(\theta^\mathcal{S}(\mathcal{S})) \\
	\quad \text{subject to} &\quad \theta^\mathcal{S}(\mathcal{S}) = \arg\min_\theta \mathcal{L}^\mathcal{S}(\theta).
\end{align}
The bi-level optimization problem incurs high computational costs and energy wastage. Therefore, it is crucial to explore approaches to streamline computation.
\subsection{Matching Gradient Based Method}
In light of the concerns regarding the memory and time complexity associated with unrolling the computational graph in meta-learning, Zhao {\it et al.} \cite{b28} introduce the matching gradient method. This method is based on the cosine similarity distance. The researchers randomly select a pair of synthetic and real batches, denoted as $\mathcal{S}_c$ and $\mathcal{T}_c$ respectively, from the datasets $\mathcal{S}$ and $\mathcal{T}$. Here, $c$ refers to the classes. In each iteration, the synthetic data for each class is updated independently. The formulation of this method is defined as follows:
\begin{align}
	\mathcal{D}(\mathcal{S},\mathcal{T},\theta) 
	&= \sum_{c=0}^{C-1} d (\nabla l(\mathcal{S}_c;\theta),\nabla l(\mathcal{T}_c;\theta)),\\
	d(\mathbf{A},\mathbf{B}) 
	&= \sum_{i=1}^{L}\sum_{j=1}^{O_i}\left(1 - \frac{\mathbf{A_j^{(i)}\cdot \mathbf{B}_j^{(i)}}}{\Vert \mathbf{A}_j^{(i)}\Vert\Vert\mathbf{B}_j^{(i)}\Vert}\right),
\end{align}
where $d(\cdot)$ represents the cosine similarity distance, $C$ signifies the total number of classes, $L$ denotes the number of layers in the neural networks, and $O_i$ signifies the output channels of the $i$th layer. 

Nevertheless, this method still necessitates considerable computational resources due to its expensive bi-level optimization problem.
\subsection{Matching Distribution Based Method}
To enhance the efficiency of the optimization process, Zhao {\it et al.} \cite{b11} propose a matching distribution method based on the Euclidean distance. They employ the Maximum Mean Discrepancy (MMD) measurement metric to match the distribution of $\theta$ between the model trained on $\mathcal{T}$ and $\mathcal{S}$. The objective function is formulated as follows:
\begin{equation}
	\mathcal{D}(\mathcal{S},\mathcal{T};\theta) = \sum_{c=0}^{C-1}\Vert \phi_{\theta^{\mathcal{S}}_c}(\tilde{x}) - \phi_{\theta^{\mathcal{T}}_c}(x) \Vert,
\end{equation}
where $\phi_{\theta^{\mathcal{S}}_c}(\tilde{x}) = \frac{1}{\mathcal{S}_c}\sum_{i=1}^{\mathcal{S}_c}f_{\theta}(\tilde{x}_i)$, and $\phi_{\theta^{\mathcal{T}}_c}(x) = \frac{1}{\mathcal{T}_c}\sum_{i=1}^{\mathcal{T}_c}f_{\theta}(x_i)$. $\mathcal{S}_c$ and $\mathcal{T}_c$ are the number of samples for the $c$th class in synthetic and real datasets respectively.
\section{Experiment}
\label{app:experiment}
\subsection{Experimental Setup}
\label{app:experiment:setup}
The performance of dataset distillation is mainly evaluated on the classification task. We follow typical settings in the area of dataset distillation, such as those outlined in DC-bench \cite{b55}, DM \cite{b11}, and IDM \cite{b56}.
\paragraph{Dataset}
We evaluate the effectiveness of our method through experiments conducted on widely-used dataset distillation benchmarks, including the MNIST \cite{b58}, Fashion-MNIST \cite{b60}, SVHN \cite{b61}, CIFAR-10, and CIFAR-100 \cite{b52}, as well as TinyImageNet \cite{b53}. The details of the datasets used in our experiments are as follows:
\begin{itemize}
    \item \textbf{MNIST \cite{b58}} comprises 60,000 training images and 10,000 testing images of grayscale handwritten digits ranging from 0 to 9. It consists of 10 classes, and each image is 28 $\times$ 28 pixels in size.
    \item \textbf{Fashion-MNIST \cite{b60}} consists of 10 classes of grayscale fashion items. The training set contains 60,000 images, and the test set contains 10,000 images. Each image is also in a 28 $\times$ 28 pixel format.
    \item \textbf{SVHN \cite{b61}} contains 73,257 training images and 26,032 test images of house numbers captured from Google Street View. It includes digit sequences ranging from 0 to 9, with each image being 32 $\times$ 32 pixels in size.
    \item \textbf{CIFAR-10 \cite{b52}} contains 60,000 32 $\times$ 32 color images distributed across 10 different classes, with 6,000 images per class.
    \item \textbf{CIFAR-100 \cite{b52}} comprises 60,000 color images, each with a resolution of 32 $\times$ 32 pixels, distributed across 100 classes. Each class contains 600 images.
    \item \textbf{TinyImageNet \cite{b53}} is a subset of the ImageNet dataset, featuring 200 classes. Each class in TinyImageNet consists of 500 training images, 50 validation images, and 50 test images, all with a resolution of 64 $\times$ 64 pixels.
\end{itemize}
\subsection{Experimental Settings}
\paragraph{Networks Architectures}
We employed the ConvNet architecture \cite{b54} to conduct dataset distillation in our experiment, adopting the approach employed in prior research \cite{b11, b34}. The ConvNet comprises three identical convolutional blocks and a linear classifier. Each block is composed of a convolutional layer with 128 kernels of size 3 $\times$ 3, instance normalization, ReLU activation, and average pooling with a stride of 2 and a size of 3 $\times$ 3. The architecture settings are consistent with those described in DC-bench \cite{b55}.
\paragraph{Evaluation Protocol}
The evaluation protocol follows the DC-bench protocol. Synthetic images are generated using 1, 10, and 50 images per class (IPC) from six benchmark datasets: MNIST, F-MNIST, SVHN, CIFAR-10/100, and TinyImageNet. To assess the effectiveness of our approach, we train a model using the generated synthetic images and measure its performance on the original test images, following the model sampling strategy \cite{b56}. Additionally, all methods employ the default data augmentation strategies provided by the authors for evaluating distillation performance. For fair comparisons in generalization evaluation, we incorporate DSA \cite{b33} data augmentation during the evaluation model training process. We report the mean accuracy of 5 runs, where the models are randomly initialized and trained for 1000 epochs, using the condensed set as the evaluation metric.
\subsection{Implementation Details}
We utilize the implementation of DM \cite{b11} and IDM \cite{b56} as a guide for setting most of the hyperparameters in our approach. To generate synthetic images, we employed the stochastic gradient descent (SGD) optimizer with a learning rate of 0.2 and a momentum of 0.5 to train synthetic datasets containing 1, 10, and 50 IPCs. For training the model, we adopted the same SGD optimizer setting with a learning rate of 0.01, momentum of 0.9, and weight decay of 0.0005. The hyperparameter $\lambda$ in $\mathcal{L}_{\text{TOTAL}} = \mathcal{L}_{\text{LH}} + \lambda \mathcal{L}_{\text{TV}} + (1-\lambda)\mathcal{L}_{\text{CLIP}}$ is set to 0.8, and the batch size is set to 256. Following the approach outlined in \cite{b33}, we employ a differentiable augmentation strategy for learning and evaluating the synthetic set. Since our approach is plug-in, we adhere to all the experimental settings (except for loss functions) of the comparison experiments and only incorporate our own modules into theirs. We conduct all experiments on clusters of NVIDIA RTX 4090 GPUs for generating synthetic datasets and one NVIDIA Tesla V100 GPU for visualizations.
\subsection{Comparison with Previous Methods}
\label{app:experiment:comparison}
We compare our proposed BACON with 10 previous methods, as listed in Table \ref{tab1} of the main content. These methods include two core-set selection methods and eight dataset distillation methods. In our comparison, we considered Random \cite{b59} and Herding \cite{b41} as coreset selection methods. Random involves randomly sampling initial synthetic images from the original dataset, while Herding selects initial synthetic images from the original dataset that are closest to the clustering center for each class. For dataset condensation methods, we included two relatively early works, DD \cite{b22} and LD \cite{b57}, as well as five advanced methods proposed later: DC \cite{b28}, DSA \cite{b33}, DCC \cite{b62}, CAFE \cite{b32}, and DM \cite{b11}. DM \cite{b11} was further improved to IDM \cite{b56}, which achieves significant performance by matching the distribution in the DD task. We choose IDM as the baseline method to evaluate the effectiveness of the proposed BACON framework in our experiments. Below are the details of these previous methods, along with the notations introduced in Appendix \ref{app:extended background}:
\paragraph{Coreset Selection}
\begin{itemize}
    \item \textbf{Random \cite{b59}} involves randomly sampling initial images from the original dataset $\mathcal{T}$ as condensed dataset $\mathcal{S}$.
    \item \textbf{Herding \cite{b41}} selects initial synthetic dataset $\mathcal{S}$ from the original dataset $\mathcal{T}$ that are closest to the clustering center for each class.
\end{itemize}
\paragraph{Dataset Distillation}

\begin{itemize}
    \item \textbf{DD \cite{b22}} first introduces the concept of Dataset Distillation (DD) and formulates the problem as a bi-level optimization. 
    \item \textbf{LD \cite{b57}} introduces a more robust and flexible meta-learning algorithm for DD, along with an effective first-order strategy utilizing convex optimization layers.
    \item \textbf{DC \cite{b28}} formulates the bi-level optimization by addressing the gradient matching problem between the gradients of deep neural network weights trained on both the original $\mathcal{T}$ and synthetic data $\mathcal{S}$.
    \item \textbf{DSA \cite{b33}} allows for the effective utilization of data augmentation to generate more informative synthetic images, thereby enhancing the performance of networks trained with augmentations.
    \item \textbf{DCC \cite{b62}} alters the loss function, enabling better comprehension of the distinctions among classes. Additionally, it introduces a pioneering bi-level warm-up strategy to enhance the stability of the optimization process.
    \item \textbf{CAFE \cite{b32}}  presents a robust approach for aligning features extracted from real $\mathcal{T}$ and synthetic datasets $\mathcal{S}$ at multiple scales, taking into account the classification of real samples $\mathcal{T}$.
    \item \textbf{DM \cite{b11}} presents a straightforward and impactful approach for generating condensed images. This is achieved by aligning the feature distributions of synthetic $\mathcal{S}$ and original $\mathcal{T}$ training images across multiple sampled embedding spaces.
    \item \textbf{IDM \cite{b56}} presents a novel dataset condensation method that is based on distribution matching, making it both efficient and promising.
\end{itemize}
\subsection{More Details of Ablation Studies}
\label{app:experiment:ablation}
We assess the effectiveness of the proposed BACON method, integrating diverse loss components, across multiple datasets. To thoroughly analyze the influence of these loss components on the distillation performance, we conduct a series of experiments using different configurations of the IPC settings. The outcomes of our experiments are presented in Table \ref{appt3}.
\begin{table}[]
  \centering
  \caption{\textbf{Ablation study of diverse loss functions:} Evaluation of the performance of the proposed loss functions, specifically $\mathcal{L}_\text{LH}$, $\mathcal{L}_\text{TV}$, and $\mathcal{L}_\text{CLIP}$, is conducted separately as loss function components. Additionally, all experimental hyperparameters, denoted by $\lambda$, are set to $0.8$ unless otherwise specified. In this experiment, we employ CIFAR-10 and CIFAR-100 as experimental datasets with setting of IPC-1, IPC-10 and IPC-50.}
  \begin{tabular}{ccccccccc}
  \toprule
  \multirow{2}{*}{$\mathcal{L}_\text{LH}$} & \multirow{2}{*}{$\mathcal{L}_\text{TV}$} & \multirow{2}{*}{$\mathcal{L}_\text{CLIP}$} & \multicolumn{3}{c}{CIFAR-10} & \multicolumn{3}{c}{CIFAR-100} \\
                      &                     &                       & IPC-50   & IPC-10   & IPC-1  & IPC-50   & IPC-10   & IPC-1   \\ \midrule
  \cmark                   &   \xmark                  &     \xmark                  & 64.86    & 55.36    & 45.32  & 41.56    & 42.68    &    \cellcolor[HTML]{D3D3D3}25.18      \\
                      \xmark& \cmark                   &  \xmark                     & 69.96    & 61.9     & \cellcolor[HTML]{D3D3D3}45.82  & 49.37    & 46.24    & 23.56        \\
                      \xmark&     \xmark                & \cmark                     & 55.07    & 42.51    & 34.22  & 30.69    & 27.49    & 15.22        \\ \midrule
                      \cmark                   & \cmark                   &   \xmark                    & 69.81    & 61.93    & 45.64  & \cellcolor[HTML]{D3D3D3}49.56    & 46.15    &  23.69       \\
                      \cmark                   &   \xmark                  & \cmark                     & 64.78    & 55.45    & 45.39  & 41.4     & 42.52         & 24.58        \\
                      \xmark& \cmark                   & \cmark                     & 69.76    & \cellcolor[HTML]{D3D3D3}62.27    & 45.69  & 49.34    &   \cellcolor[HTML]{D3D3D3}46.42        &  23.96       \\ \midrule
                      \cmark                   & \cmark                   & \cmark                     & \cellcolor[HTML]{D3D3D3}70.06    & 62.06    & 45.62  & 49.44    & 46.15    & 23.68   \\ \bottomrule
  \end{tabular}
  \label{appt3}
  \end{table}
\paragraph{Analysis} In the context of the CIFAR-10 dataset with IPC-50, BACON demonstrates superior performance by employing three distinct loss components, as detailed in the main body. However, as IPC numbers diminish, BACON utilizing total variance (TV) loss and CLIP loss emerges as the top performer at an IPC setting of 10. Conversely, under an IPC setting of 1, BACON with solely TV loss excels. Turning to the CIFAR-100 dataset, optimal performance is achieved by BACON employing likelihood (LH) loss and TV loss with the IPC-50 setting. When IPC numbers decrease, BACON with TV loss and CLIP loss achieves the highest performance. With the IPC-1 setting on the CIFAR-100 dataset, BACON with only LH loss outperforms others.
\subsection{Visualization}
\label{app:experiment:visualization}
\begin{figure}
	\centering
	\includegraphics[width=1\linewidth]{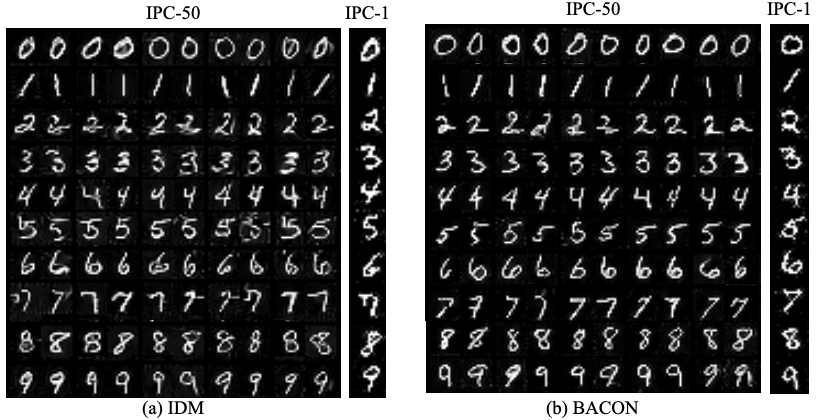}
	\caption{\textbf{Visualization of BACON and IDM on the MNIST dataset: } (a) is IDM condensed to IPC-50 and IPC-1. (b) is BACON condensed to IPC-50 and IPC-1.}
	\label{fig7}
  \end{figure}

\begin{figure}
	\centering
	\includegraphics[width=1\linewidth]{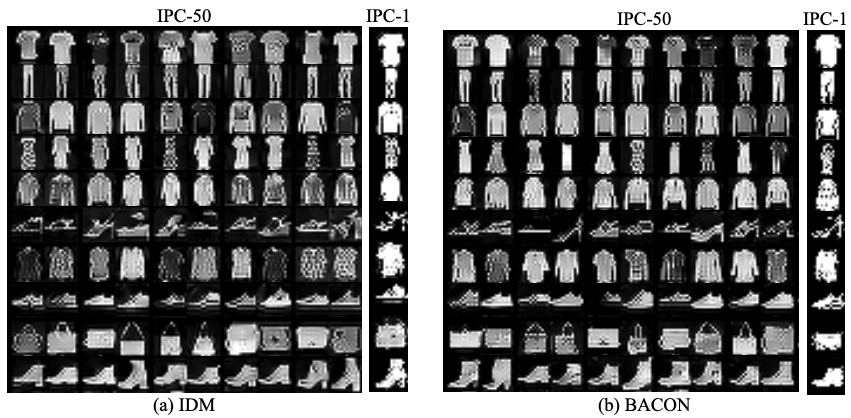}
	\caption{\textbf{Visualization of BACON and IDM on the Fashion-MNIST dataset: } (a) is IDM condensed to IPC-50 and IPC-1. (b) is BACON condensed to IPC-50 and IPC-1.}
	\label{fig8}
  \end{figure}

\begin{figure}
	\centering
	\includegraphics[width=1\linewidth]{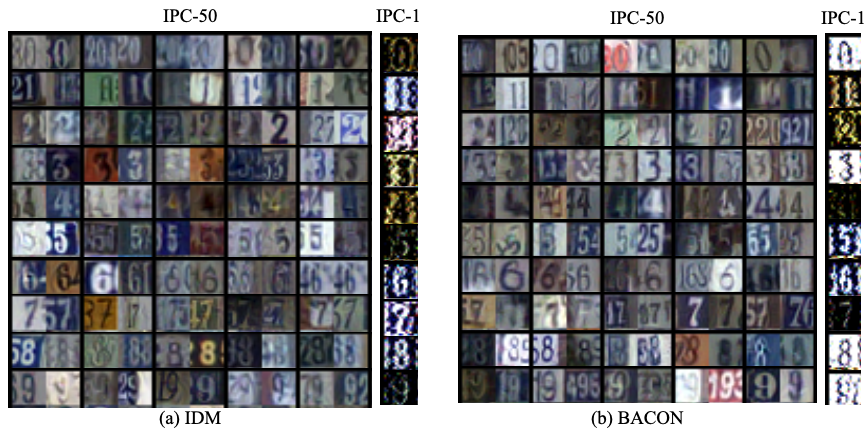}
	\caption{\textbf{Visualization of BACON and IDM on the SVHN dataset: } (a) is IDM condensed to IPC-50 and IPC-1. (b) is BACON condensed to IPC-50 and IPC-1.}
	\label{fig9}
  \end{figure}

\begin{figure}
	\centering
	\includegraphics[width=1\linewidth]{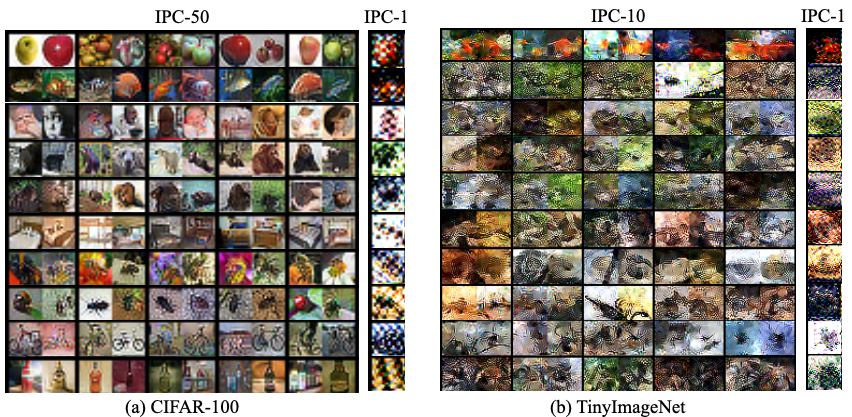}
	\caption{\textbf{Visualization of BACON on the CIFAR-100 and TinyImageNet datasets: } (a) is BACON condensed to IPC-50 and IPC-1 on the CIFAR-100 dataset. (b) is BACON condensed to IPC-10 and IPC-1 on the TinyImageNet.}
	\label{fig10}
  \end{figure}

To demonstrate the effects of the distilled images in a more intuitive manner, we conducted a comparative analysis of the distillation results obtained using the proposed BACON method and IDM \cite{b56} on extensive datasets, namely MNIST, Fashion-MNIST, and SVHN. The visual comparisons are depicted in Figure \ref{fig7}, \ref{fig8}, and \ref{fig9}. Moreover, we present additional visualizations of the distilled images obtained from CIFAR-100 and TinyImageNet in Figure \ref{fig10}.
\section{Broader Impacts}
\label{app:impact}
The introduction of BACON, a new framework for Dataset Distillation (DD), brings promising advantages by reducing storage costs and training expenses while maintaining performance on test sets. This breakthrough could make large datasets and models more accessible, opening up opportunities for innovation in fields like healthcare, education, and climate science. However, it is crucial to address potential biases and ethical concerns in the distillation process to ensure fairness and accountability. Furthermore, BACON's theoretical groundwork not only deepens our understanding but also sets the stage for further progress in machine learning research, emphasizing the need for responsible development and deployment of distillation methods.
\section{Limitations and Future Works}
\label{app:limitation}
Although the proposed BACON method effectively enhances model performance on synthetic datasets, its efficacy declines as the IPC setting of synthetic images decreases. Additionally, as image resolution increases, computational costs escalate accordingly. Therefore, our future work will focus on addressing these limitations. Specifically, we aim to enhance BACON's performance in high-resolution image scenarios while simultaneously improving computational efficiency to reduce energy overhead.

\clearpage

\end{document}